\documentclass{article}

\usepackage{graphicx}
\usepackage{graphics}
\usepackage{amsfonts}
\newcommand{\R}{\mathbb{R}}
\newcommand{\Z}{\mathbb{Z}}

\begin{document}
\title{On Ultrametric Algorithmic Information}
\author{Fionn Murtagh \\
Department of Computer Science,
Royal Holloway, \\
University of London, Egham TW20 0EX, England \\
 (fmurtagh symbolat acm dot org)}

\maketitle

\begin{abstract}
How best to quantify the information of an object, whether natural or 
artifact, is a problem of wide interest.  A related problem is the 
computability of an object.  We present practical examples of a new way to 
address this problem.  By giving an appropriate representation to our 
objects, based on a hierarchical coding of information, we exemplify how 
it is remarkably easy to compute complex objects.  Our algorithmic 
complexity is related to the length of the class of objects, rather than 
to the length of the object.  
\end{abstract}


\section{Introduction}

Brooks \cite{brooks} asserted as a great challenge for 
contemporary computer science and information theory: 
``Shannon and Weaver performed an inestimable service by giving 
us a definition of information and a metric for information as 
communicated from place to place. We have no theory however that 
gives us a metric for the information embodied in structure ... this 
is the most fundamental gap in the theoretical underpinning of 
information and computer science.''  

The notion of ultrametric information was introduced by \cite{khrenn}, both
to handle interactive as opposed to static information, and by taking 
a dynamic view of information, with analogies to metric or Kolmogorov-Sinai
information.  Here we pursue a view of algorithmic or computational
information, which is extended to account for an ultrametric
embedding of the object that is  computed.  

Shannon information is oriented towards communication.
While Shannon information is based on the freedom of choice that is
possible when transmitting a message, Kolmogorov information, or
algorithmic information, is a measure of the information content of
individual objects.  
The Kolmogorov complexity of a string is the size of the shortest
program in bits that computes the string.  It is concerned therefore
with strings, and furthermore (finite or infinite) binary strings.
An object, expressed as a binary string, has complexity which is its
shortest string description, because this also defines the shortest
program, or decision tree, to compute it.
The shortest effective description length has become known as
Kolmogorov complexity, even if precedence may be due to Solomonoff
(\cite{vitanyi}, p.\ 90).
From Solomonoff's work on the ``algorithmic theory of descriptions''
has  come the minimum description length, or MDL, principle as a computable
and practical information measure \cite{barron98,wallace99,rissanen06}

We approach this problem of expressing information and computability,
relating to complexity and generation, respectively, in a new way.  
A key role is played by representation, i.e.,  object or data encoding.  We
need both to consider carefully the data description related to the
observing of the object; and the display associated with the data description.
These two issues amount to, respectively, the mapping of the object to data, 
and data to object.
There is enormous latitude for representation.  We must choose
expeditiously, based on our objectives, which may include interpretation
of the data or the event or phenomenon. 

Our work builds on \cite{murtaghjoc,murtaghca} in the following ways.  
Such 
work points to the crucial role played by data encoding, or representation,
for many purposes (including search and display).  In \cite{murtaghjoc} it is 
shown how {\em if} we have an ultrametric embedding of our data -- otherwise 
expressed, a hierarchical or tree
structuring of our data -- then it is possible for 
search operations to be carried out in constant, or $O(1)$, time. 
In this article, we also presuppose a given ultrametric embedding (or hierarchical
structuring) of our data.   In general terms we are 
dealing with $n$ objects characterized by $m$ attributes.  Classically,
a complete description of an object by means of its attributes leads to an
expression for the object's complexity that is defined 
from the set of 
its $m$ attributes.  Given the ultrametric embedding, we look instead at the 
object's complexity in terms 
that are relative to the population of $n$ objects.  
If the hierarchy is a meaningful one, e.g.\ expressing biological 
reproduction, then we have a new perspective on the computability of an object.

Section \ref{sect2} provides background on an important tool used in subsequent
sections, the Haar wavelet transform carried out {\em on} 
a hierarchy.  It allows us
to go well beyond a hierarchy as just a display device or visualization, and 
instead to carry out operations on the hierarchy, expressing operations in an 
ultrametric space. 
We set the scene for later parts of this article 
through a discussion of the stepwise
approximation scheme that we can establish, for various objects, and 
that defines the Haar wavelet transform, in this case, of a dendrogram.  
(A dendrogram is the term used for the particular tree, discussed in the next
section, that is induced on, or determined from, object/attribute data.  
In this article, our use of the term ``hierarchy'' is always as a synomym 
for these.)

In section \ref{sect6} we consider a hugely simplified face recognition
case study.  Once we presuppose a representation or encoding of a face, 
then any given face is generated by very simple calculations on faces.
We link this work with some recent directions of study in the psychology 
literature of human recognition behavior. 

In section \ref{sectnew4} we use  a simple case study of a set of concepts, 
and show how each is computed or generated from others among these concepts, 
and/or a superset of nouns.  This study 
is complemented by the analysis of texts 
or documents.  

In dealing with faces and with texts, we have carefully selected a range
of case studies to exemplify a new approach to computability, in the sense of
generation of an object and, related to this, the inherent 
complexity of an object.  

In summarizing and concluding, sections \ref{sect55} and \ref{sect66} 
provide further discussion on our approach. 

\section{Wavelet Transform of a Set of Points Endowed with an Ultrametric}
\label{sect2}

\subsection{Description Using an Example}

A wavelet transform is a decomposition of an object, typically an image or
signal, into an ordered set of 
{\em detail} ``versions'' of the data, and an 
overall {\em smooth} \cite{starck}.  
From the details, with the smooth, the data can be 
exactly reconstructed.  In the case of the Haar wavelet transform, the details 
and the smooth are defined from, respectively, differences and sums.  We will 
see how this works using a concrete example.

Extending the wavelet transform to ultrametric topologies has been 
carried out, e.g., by \cite{khrennkoz05,khrennkoz}.  
The wavelet transform has been 
traditionally used for image and signal processing, based on functions in
 Hilbert space.  In \cite{denhaar} we showed, with a wide range of examples
and case studies, how this transform can be easily implemented {\em on} 
tree structured data.  Without loss of generality, we assume that our tree
is binary, rank ordered, rooted, and, for practical application, labeled.
Such a tree is often referred to as a dendrogram.  The tree distance is 
an ultrametric and, reciprocally, we endow a data set with an ultrametric by
structuring it as a tree.  



As a small data set consider the first 8 observations in the very widely 
used Fisher iris data \cite{fish}.
Fisher used this data, taken from \cite{anderson}, 
to introduce the discriminant analysis method that bears his name.
By range-normalizing (i.e., subtracting the minimum value of each variable, and
dividing by the range) in Table \ref{table5}, we obtain Table \ref{table5b}.  

\begin{figure*}
\begin{center}
\includegraphics[width=14cm]{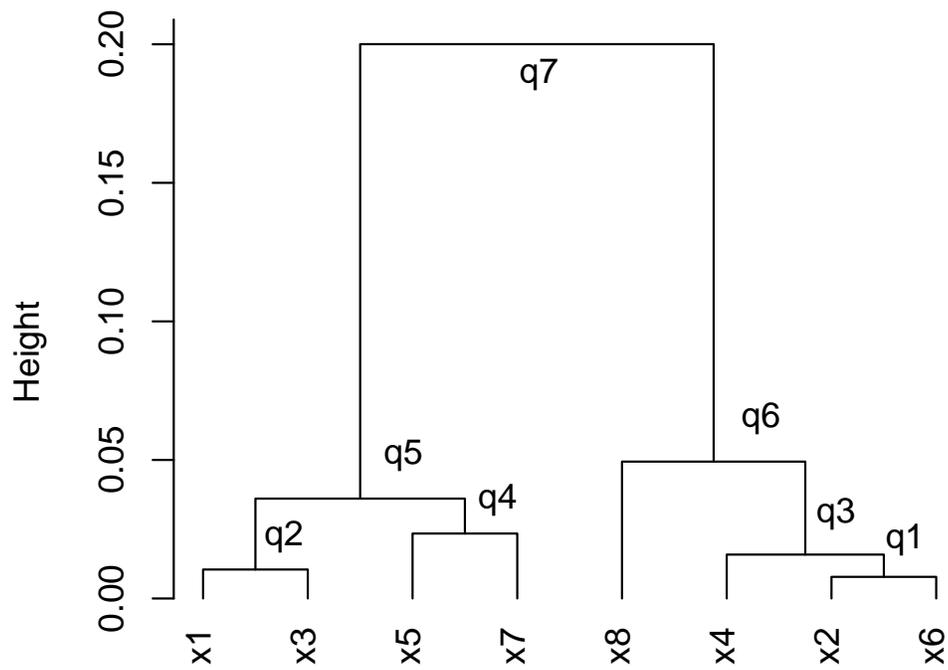}
\end{center}
\caption{Ward minimum variance hierarchy of the data shown in Table
\ref{table5b}.  The clusters are labeled $q_1$, $q_2$, etc.}
\label{fig5b}
\end{figure*}

The minimum variance
or Ward agglomerative clustering hierarchy was built (with constant
weights on the observations), and is shown in Figure \ref{fig5b}.
The minimum variance agglomeration criterion, with Euclidean
distance, is used to induce the hierarchy on the given data.  We could use
some other agglomerative criterion.  However the minimum variance one
leads to more balanced dendrograms \cite{mur84,murtaghca}, with knock-on
implications for computational requirements for average time tree traversal.

\begin{figure*}
\begin{center}
\includegraphics[width=14cm]{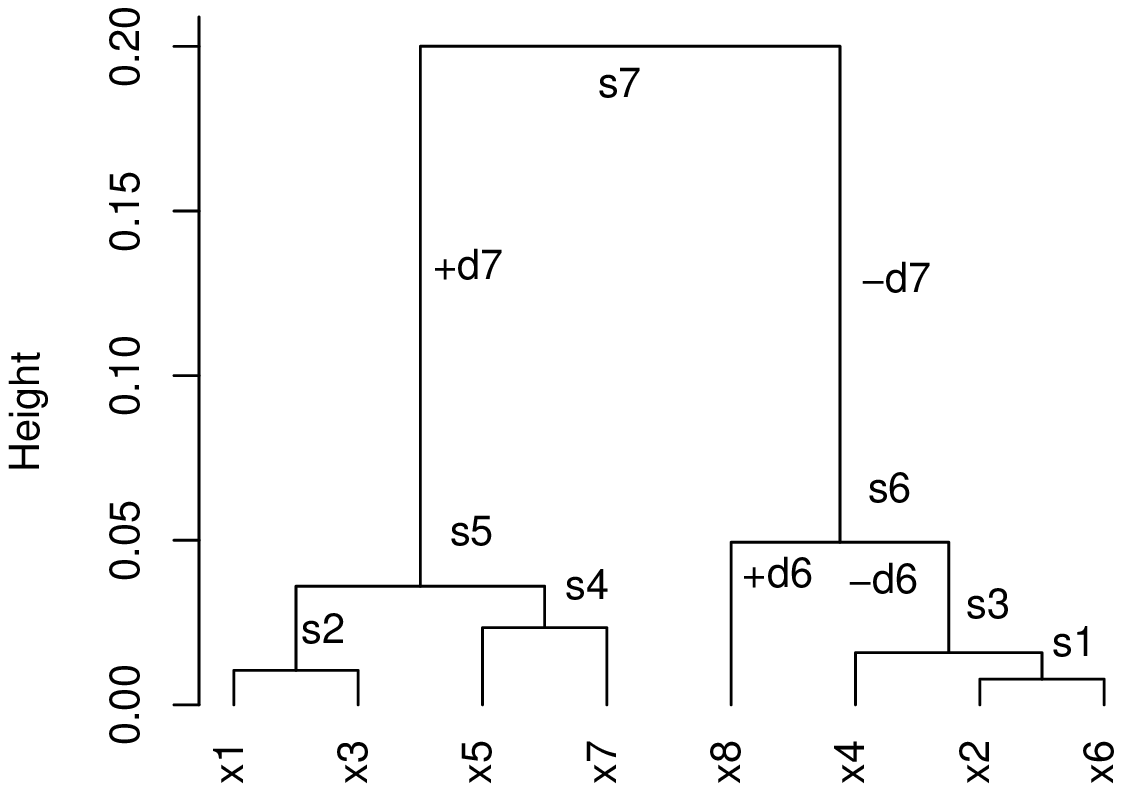}
\end{center}
\caption{As Figure \ref{fig5b}, with {\em smooths}, s7, s6, etc.\ shown,
and with some of the {\em detail} vectors, d7, d6.  Details (shown) +d7 and $-$d7 are 
associated with offspring branches of node s7.  Details (shown) +d6 and $-$d6 are
associated with offspring branches of node s6.  Details (again shown) +d5 and $-$d5 are
associated with offspring branches of node s5. The situation is analogous to this 
(although not shown) for nodes s2, s5, s4, s3 and s1.}
\label{fig5c}
\end{figure*}

From input Table \ref{table5b} and the dendrogram of Figure \ref{fig5b}, 
we carry out the wavelet transform.  The transform is shown in Table 
\ref{table6b}, and is also displayed in Figure \ref{fig5c}.

Note that in Table \ref{table6b} it is entirely appropriate that at more
smooth levels (i.e., as we proceed through levels d1, d2, $\dots$, d6, d7)
the values become more ``fractionated'' (i.e., there are
more values after the decimal point).
Each detail
signal is of dimension $m = 4$ where $m$ is the same 
dimensionality as the given, input, 
data.  The smooth signal is of dimensionality $m$ also.  The number of
detail or wavelet signal levels is given by the number of levels in the
labeled, ranked hierarchy, i.e.\ $n-1$: cf.\ the columns in Table 
\ref{table6b} labeled, for details, $d7, d6, \dots$.

\begin{table*}
\caption{First 8 observations of Fisher's iris data.  L and W
refer to length and width.} 
\label{table5}
\begin{tabular}{rrrrr}\hline
    &   Sepal.L   &  Sepal.W    &  Petal.L  &   Petal.W \\\hline
1   &       5.1   &      3.5    &      1.4  &       0.2 \\
2   &       4.9   &      3.0    &      1.4  &       0.2 \\
3   &       4.7   &      3.2    &      1.3  &       0.2 \\
4   &       4.6   &      3.1    &      1.5  &       0.2 \\
5   &       5.0   &      3.6    &      1.4  &       0.2 \\
6   &       5.4   &      3.9    &      1.7  &       0.4 \\
7   &       4.6   &      3.4    &      1.4  &       0.3 \\
8   &       5.0   &      3.4    &      1.5  &       0.2 \\\hline
\end{tabular}
\end{table*}

\begin{table*}
\caption{First 8 observations of Fisher's iris data.  L and W
refer to length and width.  Values are range-normalized (in each column:
minimum subtracted, and divided by range).}
\label{table5b}
\begin{tabular}{rrrrr} \hline
    &   Sepal.L   &  Sepal.W    &  Petal.L  &   Petal.W \\ \hline
1   &       0.625   &    0.5556    &     0.25  &       0.0 \\
2   &       0.275   &    0.0   &      0.25  &       0.0 \\
3   &       0.125 &      0.2222    &   0.0  &       0.0 \\
4   &       0.0   &      0.1111    &   0.5  &       0.0 \\
5   &       0.5   &      0.6667   &     0.25  &       0.0 \\
6   &       1.0   &      1.0    &      1.0  &       1.0 \\
7   &       0.0   &      0.4444    &     0.25  &       0.5 \\
8   &       0.5   &      0.4444    &     0.5  &       0.0 \\ \hline
\end{tabular}
\end{table*}



\begin{table*}
\caption{The hierarchical Haar wavelet transform resulting from the
hierarchy of Figure \ref{fig5b}, built on the data of Table \ref{table5b}.
Last data {\em smooth}: s7; levels of {\em detail} from top to bottom
(presented left to right), d7, d6, $\dots$, d2, d1.  We used the convention
that the left subnode has a positive {\em detail}, and the right subnode has a
negative {\em detail}. (Data precision here to 4 decimal places.)}
\label{table6b}
\begin{tabular}{rrrrrrrrr} \hline
        &     s7  &     d7    &    d6  &   d5  &   d4   &  d3  &  d2 &    d1 \\
\hline
Sepal.L &  0.3672 & $-0.0547$ &  0.0781 &  0.0625 &  0.25 & $-0.3438$ & 0.25 & $-0.3125$ \\
Sepal.W & 0.4236  & 0.0486  & 0.0694 & $-0.0833$  & 0.1111 & $-0.1944$ & 0.1667 & $-0.5$ \\
Petal.L & 0.3594 & $-0.1719$ & $-0.0313$ & $-0.0625$ &  0.0 & $-0.0625$ & 0.125 & $-0.375$ \
\\
Petal.W &  0.125 & 0  & $-0.125$ & $-0.125$ & $-0.25$ & -0.25 & 0 & $-0.5$ \\ \hline
\end{tabular}
\end{table*}

To summarize, we begin typically with an object set, 
each object having values on 
an attribute set.  From this, a hierarchy of the objects 
is created.  Then this 
hierarchy is further processed.  We get a set of details and a smooth vector,
such that they suffice for reconstruction of the input data.  The hierarchy 
provides a ``key'' for us to recreate the input data.  The total number of 
values in the dendrogram wavelet transformed data is precisely the same 
as the number of values in the input data.  

\subsection{Representation of an Object as a Chain of Successively Finer
Approximations}
\label{sect3}

From the wavelet transformed hierarchy  
we can read off that, say, $x_1 = d_2 + d_5 + d_7 + s_7$:
cf.\ Figure \ref{fig5c}.   Or $x_8 = d_6 - d_7 + s_7$.
These relationships use the appropriate vectors 
shown (as column vectors) in Table \ref{table6b}.  
Such relationships furnish the definitions used by
 the inverse wavelet transform, i.e.\ the recreation of the input data
from the transformed data.

Thus, the Haar dendrogram wavelet transform gives us 
 an additive decomposition of a given observation
(say, $x_1$) in terms of a degrading approximation, with a variable number
of terms in the decomposition.  The objects, or observations, are
those things which we are analyzing and on which we have (i) induced a 
hierarchical clustering, and (ii) further processed the hierarchical 
clustering in such a way that we can derive the Haar decomposition.  
In this section we will look at how this allows us to consider each object
as a limit point. 
Our interest lies in our object set, characterized by a set of data, 
 as a set of limit or fixed points.  


Using notation from domain theory (see, e.g., \cite{edalat97}) we write:

\begin{equation}
 s_7 \sqsubseteq s_7 + d_7 \sqsubseteq s_7 + d_7 + d_5
\sqsubseteq s_7 + d_7 + d_5 + d_2
\label{eqn5}
\end{equation}

The relation $a \sqsubseteq b$ is read: $a$ is an approximation to $b$,
or $b$ gives more information than $a$.  (Edalat \cite{edalat03} 
discusses examples.)
Just rewriting the very last, or rightmost, term in relation (\ref{eqn5})
gives:

\begin{equation}
s_7 \sqsubseteq s_7 + d_7 \sqsubseteq s_7 + d_7 + d_5
\sqsubseteq x_1
\label{eqn5b}
\end{equation}

Every one of our observation vectors (here, e.g., $x_1$) can be
increasingly well approximated by a {\em chain} of the sort shown in
relations (\ref{eqn5}) or (\ref{eqn5b}), 
starting with a least element ($s_7$; more generally,
for $n$ observation vectors, $s_{n-1}$).  The observation vector itself (e.g.,
$x_1$) is a least upper bound (lub) or supremum (sup), denoted
$\sqcup$ in domain theory, of this chain.  Since every observation vector
has an associated chain, 
every chain
has a lub.  The elements of 
the ``rolled down''
tree, $s_7$, $s_7 + d_7$ and $s_7 - d_7$,  $s_7 + d_7 +d_5$ and
$s_7 + d_7 - d_5$, and so on, are clearly representable as a binary rooted
tree, and the elements themselves comprise a partially ordered set (or
poset).  A {\em complete partial order} or {\em cpo} or {\em domain} is a
poset with least element, and such that every chain has a lub.  Cpos generalize
complete lattices: see \cite{davey} for lattices, domains, and their 
use in fixpoint applications.  


\subsection{Approximation Chain using a Hierarchy}

An alternative, although closely related, structure with which domains
are endowed is that of spherically complete ultrametric spaces.   The
motivation comes from logic programming, where non-monotonicity may well
be relevant (this arises, for example, with the negation operator).  Trees
can easily represent positive and negative assertions.  The general notion
of convergence, now, is related to {\em spherical completeness} 
(\cite{schikhof,hitzler}; see also \cite{khrenn}, Theorem 4.1).  
If we have any set of embedded clusters,
or any chain, $q_k$, then the condition that such a chain be non-empty,
$\bigcap_k q_k \neq \emptyset$, means that this ultrametric space is
non-empty.  This gives us both a concept of completeness, and also a
fixed point which is associated with the ``best approximation'' of the
chain.

Consider our space of observations, $X = \{ x_i | i \in I \}$.  The
hierarchy, $H$, or binary rooted tree, defines an ultrametric space.  For
each observation $x_i$, by considering the chain from root cluster to
the observation, we see that $H$ is a spherically complete ultrametric
space.

\subsection{Mapping of Spherically Complete Space into Dendrogram Wavelet
Transform Space}

Consider analysis of the set of observations, $\{ x_i \in X \subset 
\R^m \}$.  Through use of any hierarchical clustering (subject to 
being binary, a sufficient condition for which is that a pairwise 
agglomerative algorithm was used to construct the hierarchy), followed by 
the Haar wavelet transform of the dendrogram, we have an approximation 
chain for each $x_i \in X$.  This approximation chain is defined in terms of
embedded sets.  Let $n = \mbox{card }  X$, the cardinality of the set $X$.  
Our Haar dendrogram wavelet transform
allows us to associate the set $\{ \nu_j | 1 \leq j \leq n-1 \} \subset 
\R^m$ with the chains, as seen in section \ref{sect3}.  

We have two associated vantage points on the generation of observation
$i, \forall i$: 
the set of embedded sets in the approximation chain starting always with the 
entire observation set indexed by the set $I$, and ending with the 
singleton observation; or
the global smooth in the Haar transform, that we will call $\nu_{n-1}$, 
running through all details 
$\nu_j$ on the path, such that an additive combination of 
path members increasingly approximates the vector $x_i$ that corresponds to 
observation $i$.  Our two associated views are, respectively, a set of 
sets; or a set of vectors in $\R^m$.  We recall that $m$ is the 
dimensionality of the embedding space of our observations.  Our two 
associated views of the (re)generation of an observation  both 
rest on the hierarchical or tree structuring of our data.

\section{Generating Faces}
\label{sect6}



\subsection{A Simplified Model of Face Generation}

Consider a very simplified model of face recognition, providing a 
``toy problem'', from which we will draw some important conclusions.  
Representation or encoding ``takes the strain'' of 
our approach, so we need to have that addressed as a matter of priority.  
For the link with human neural encoding of faces, \cite{treves} is a 
useful starting point.  A ``perceptual face space'' is at 
issue in \cite{treves}
and this author proceeds to point to limits of Euclidean embedding of 
perceptual face spaces, and instead proposes arguments in favor of 
ultrametric embedding.  Therefore \cite{treves} is a very useful 
prolegomenon for our current work.  

We codify our simplified and stylized faces in an analogous way to 
the encoding often used in the processing of real faces \cite{young}.
We use \cite{wolf} and associated software in R, and also the results
presented here that are based on an implementation of Chernoff
\cite{chernoff} in S-Plus.   We will scale  data such that all 
attributes are in the interval 0, 1.  We use 15 attributes for a face,
given as follows: 1 -- area  of  face;  2 -- shape  of
       face;  3 -- length  of  nose; 4 -- location of mouth; 5 -- curve of
       smile;  6 -- width  of  mouth;  7,  8,  9,  10,  11 -- location,
       separation, angle, shape and width of eyes; 12 -- location of
       pupil; 13, 14, 15 -- location, angle and width of eyebrow.

\begin{figure*}
\begin{center}
\includegraphics[width=14cm]{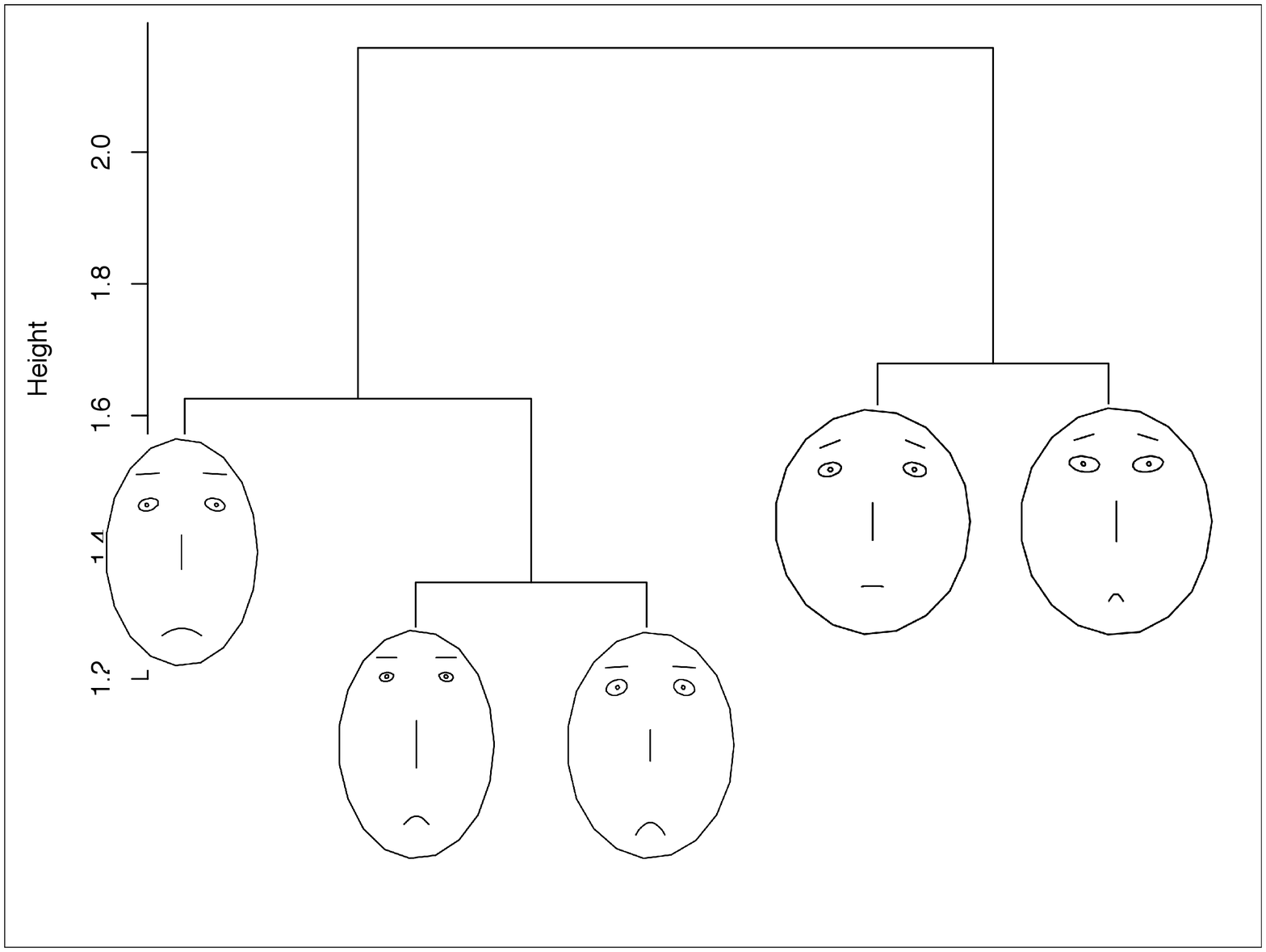}
\end{center}
\caption{Five randomly generated Chernoff faces which were then hierarchically
clustered.  The depictions of faces are defined from attribute vectors.  The 
actual processing takes place, of course, on the numeric representation.}
\label{chern1}
\end{figure*}

\begin{figure*}
\begin{center}
\includegraphics[width=14cm]{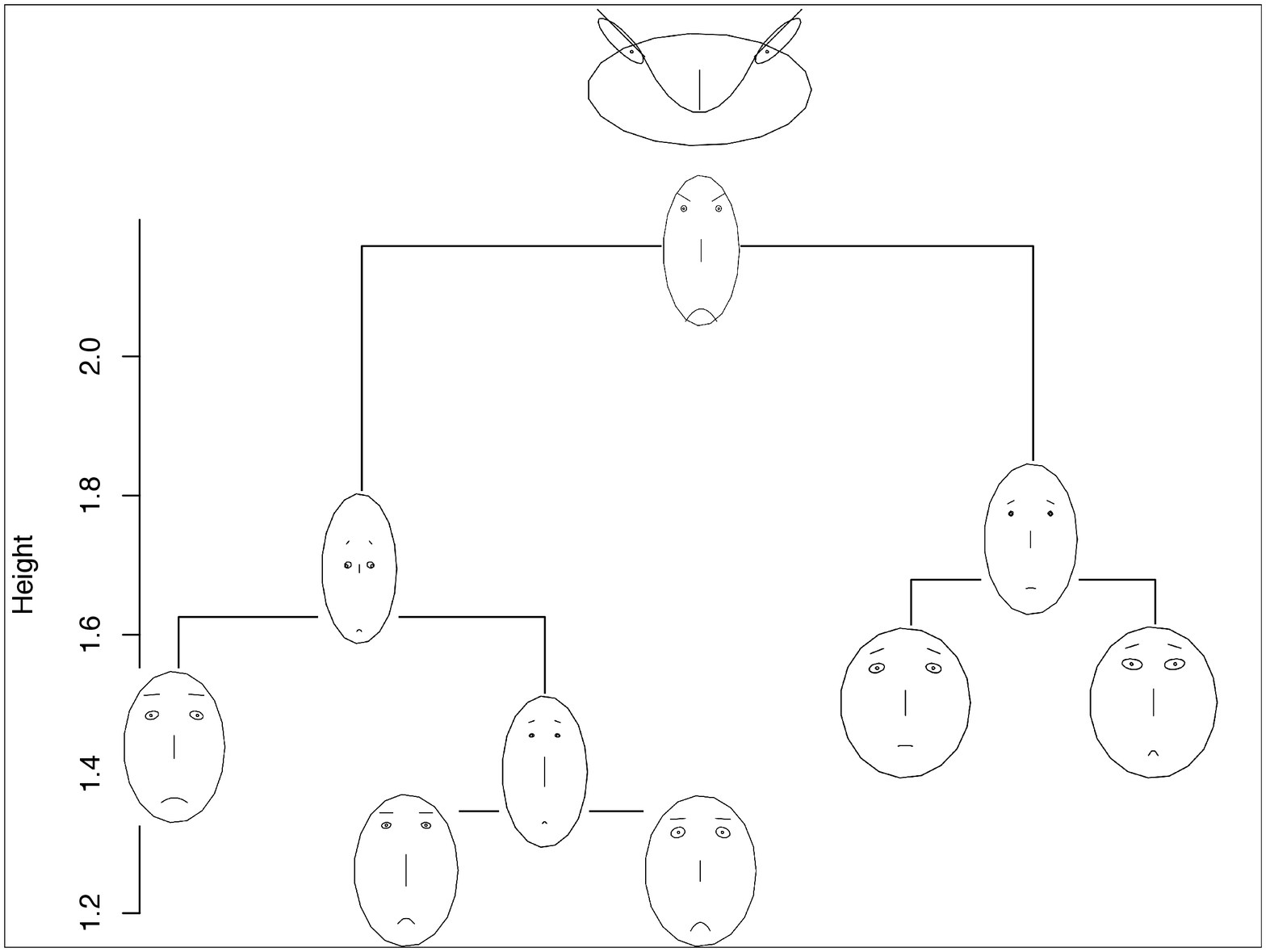}
\end{center}
\caption{The dendrogram Haar wavelet transform of the 5 faces shown 
in Figure \ref{chern1}, where the detail and smooth signals are 
displayed as faces.  The face at the very top is the final smooth.  The second 
highest face is the detail face that must be added to, or subtracted from, the 
final smooth, in order to yield the smooths at the next level down, to left and
right.  With these smooths, and plus (right offspring branch) or minus
(left offspring branch)  
the details shown here, we can proceed to the 
next levels down.  In this way, we recreate the original (input) data in a 
stepwise fashion, following the branches of the tree.  Here, we also show the 
terminal nodes (identical to Figure \ref{chern1}).}
\label{chern2}
\end{figure*}

Figure \ref{chern1} shows 5 randomly generated (uniformly on the 
15 attributes) faces.  
A hierarchical clustering (Ward minimum variance criterion
used) has been carried out in this figure.  
Then a Haar dendrogram wavelet 
transform was applied, based on a lifting scheme implementation
(described in section \ref{sect43} below).  The point 
of relevance in this implementation is that details and the smooth are 
defined from sums and differences; then in reconstructing the data, means
are used (see Table \ref{tabhaar}, to be discussed below).  
The result of the wavelet transform is shown in Figure \ref{chern2},
where detail coefficients and the smooth are depicted as faces.  

By proper combination of smooth and details (in 
Figure \ref{chern2}), each one of the faces in Figure \ref{chern1} can be 
exactly reconstructed.  Note that what we have here are mappings of data sets
onto the facial representations, which means that the data that we 
calculate with are encodings of these facial representations.  
We have a well-defined and unique procedure for (i) decomposing or
``peeling away'' the input data to yield the transformed data; and
(ii) a recomposition, allowing regeneration of the input data. 

The smooth in Figure
\ref{chern2} is the sum of all faces.  (Interestingly, the city of Sydney
has determined ``real life'' average faces, involving a great number of 
people.  These average faces  
are identical to sums, modulo scaling.  See 
\cite{fos}.)   

When the dendrogram is ``balanced'' or 
``symmetric'' \cite{mur84},   the smooth is, to within a constant, 
the (unweighted) mean object;
and the path traversed in 
the dendrogram, our ``key'' to reconstituting a face, has approximately 
$\log n$ steps on it.  

\subsection{Discussion}

As noted the overall smooth, and start point for the reconstruction of 
any object
from the hierarchically represented information, can be to within a constant 
the mean
object.  Our approach uses a hierarchy as a
``key'' to the generative mechanism for an object.  Our approach is
therefore a norm-referenced one.

In \cite{giese}, it is found that norm-referenced encoding of human 
faces is a more likely mechanism in facial recognition, compared to 
example-based encoding.  The former is with
reference to an average or norm, whereas the latter is relative to 
prototypical faces.  \cite{giese2} reinforces this: ``The main finding
was a striking tendency for neurons to show tuning that appeared 
centered about the average face''.  They suggest that norm-referencing
is helpful for making face recognition robust relative to viewing 
angle, facial expression, age, and other variable characteristics.
Finally they suggest: ``Norm-based mechanisms, having adapted to our
precise needs in face recognition, may also help explain why our face
recognition is so immediate and effortless...''.


A wide range of experimental psychology results are presented by 
\cite{dijksterhuis} to support the link between norm-referenced reasoning and 
unconscious reasoning, on the one hand, 
contrasted with the link between prototype-referenced 
reasoning and conscious thinking, on the other hand.  We will pursue some 
discussion of these links since they provide a most consistent backdrop to our 
work.  

Encoding of information is fundamental.  
``Thinking about an object implies that the representation of
that object in memory changes.''  Furthermore, 
``information acquisition'' remains crucial for either
form of thought, conscious or unconscious.  

Dijksterhuis and Nordgren \cite{dijksterhuis} point to how conscious thought 
can process between 10 and 60 bits per second.  In reading, one processes about 
45 bits per second, which corresponds to the time it takes to read a fairly short 
sentence.  However the visual system alone processes about 10 million bits per 
second.  It is concluded from this that the conscious thinking process in 
humans is very low, compared to the processing capacity of the entire human 
perception system.
 Conscious thought therefore is
both limited and limiting.  A small number of foci of interest
(``only one or two attributes'') have to
take priority.  There are inherent limits to conscious thought as a result.
 As a result of limited capacity,
``conscious thought is guided by expectancies and schemas''.
Limited capacity therefore goes hand in hand with use of stereotypes or schemas.
``... people use ... stereotypes (or schemas in general) under
circumstances of constrained processing capacity ...  [While]
this [gives rise to the conclusion] that limited processing capacity
during {\em encoding} of information leads to more schema use,
[current work proposes] that this is also true for thought
processes that occur after encoding.  ... people stereotype more
during impression formation when they think consciously compared
to when they think unconsciously.  After all, it is consciousness
that suffers from limited capacity.''

It may, Dijksterhuis and Nordgren \cite{dijksterhuis} proceed,
 be considered counter-intuitive that stereotypes are
applied in the limited capacity, conscious thought, regime.
However stereotypes may be ``activated automatically (i.e.,
unconsciously)'', but ``they are applied {\em while we
consciously think} about a person or group''.
Conscious thought is therefore more likely to (unknowingly)
attempt ``to confirm an expectancy already made''.  

On the other hand,
unconscious thought is less biased in this way, and more slowly 
integrates information.  ``Unconscious thought leads
to a {\em better organization} in memory'', arrived at through ``incubation''
of ideas and concepts.  
``The unconscious works ... aschematically, whereas consciousness
works ...  schematically''.   ``... conscious thought is more like
an architect, whereas unconscious thought behaves more like an archaeologist''.

Viewed from the perspective of the work discussed in this subsection, it can
be appreciated that our hierarchical and generative description of an object 
set is a simple model of unconscious thought.  (That it is simple is clear: 
to begin with, it is static.)   Our hierarchical and generative description of 
an object set is due to the object set being embedded in  an 
ultrametric topology.  In this framework, then, the information content is
defined from the size of the object set, and not from any given object.

\section{Generating Literary Texts}
\label{sectnew4}

\subsection{Spherically Complete Ultrametric Text Space}

The face case-study was based on normalized data, and with arbitrary and 
limitless potential for generating new faces.  Practical data analysis, 
on the other hand, often deals with a limited number of objects.   We 
will set up case studies to explore some such situations.

Consider the total literary output of an individual or
group of individuals.  As a simplified 
case-study we will use a set of 209 Grimm Brothers'
tales, in English.  We want to explore how our ``norm-based'' approach, 
based on a hierarchical structuring of the set of 209 text objects, allows us 
to consider any given tale to be generated from the average one; as opposed
to the creation of the text in some other isolated way, without 
reference to its peer texts.   

Encoding of the data is our first step.  
We took 209 tales of the Grimm Brothers (data available from 
\cite{ockerbloom}).  
  There were, in all, 280,629 words.  Story lengths were between
650 and 44,400 words.  A frequency of occurrence cross-tabulation was
formed of the 209 texts and 7443 unique words.  To handle normalization,
the $\chi^2$ distance between text {\em profiles} was used as input to
correspondence analysis, which furnished an output Euclidean embedding of
dimensionality one less (a linear dependence due to the centering) 
than min(209, 7443) (dual space relationship) \cite{murtaghca}.  The
minimum variance agglomerative hierarchical clustering of the 209
tales (identically weighted)  was carried out, using their 208-dimensional
Euclidean embedding.  The Haar wavelet transform was applied then to this
hierarchy.

\begin{figure*}
\begin{center}
\includegraphics[width=6cm]{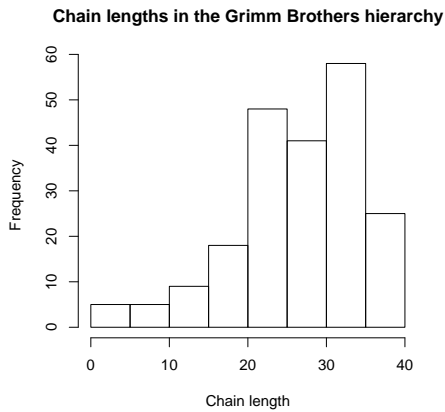}
\end{center}
\caption{Histogram of chain lengths, always from root cluster to
terminal observation, from the hierarchy constructed on the 209
Grimm Brothers tales.}
\label{fig88}
\end{figure*}

Figure \ref{fig88} shows the histogram of chain lengths, with mean
26.70, and median 28.
All chains are from root to a terminal node.  For $n$ terminals,
obviously there are $n$ chains.  The chains are derived from our
hierarchy.  It is the Haar wavelet transform that gives us an
interpretation of the chains in terms of progressively better
approximations.

A few further comments on this wavelet decomposition follow.  Consider
each chain which starts at the root cluster vector, and
ends with an observation vector.  In all cases, in this study of wavelet
transform properties, these
vectors are of dimensionality 208.
We will comment on the Grimm Brothers tales, knowing that, in this
case, we are taking each such tale as defined by its 208-valued vector.
Each tale is a point in $\R^{208}$.  This is 
somewhat of an over-simplification, evidently, since word order is not 
taken into account.  However this ``bag of words'' approach will be 
adequate as  a first  model of literary creation. 
Each chain furnishes a monotonically improving 
approximation of a Grimm Brothers tale. Furthermore the point of
departure for all tales is the same, viz.\ the vector associated with the
root node, or $s_{207}$ to use notation from section \ref{sect2}.
From this common point of departure, an approximating chain of length
maximum 40 (and of minimum 1) was found to suffice to ``create''
or ``generate'' the Grimm tale.
Thus at most 40 transitions (and at least 1 transition) were required to
create a tale from the common starting material, which to within a constant 
(and with no loss of generality), approximates the ``norm''.  
(The term ``norm'' is 
used as in psychology, not as in mathematics.) An algorithm 
to generate a tale, in this framework, is 
of worst case computational complexity linear in the number of
tales.  The more usual probabilistic perspective is where the
tale has to be assembled from its components, and this is seen to imply a
computational complexity that is linear in
the ambient dimensionality of the space used, e.g.\ a space of words.  

Our wavelet transform allows us to read off the chains that make the
ultrametric space a spherically complete one.   An 
 $O(n)$ data (re-)generation algorithm ensues, compared to a more
usual  $O(m)$ data generation algorithm.  Here, $n$ is the number of tales, 
and $m$ is the number of words used to characterize them.  The
importance of our result is when $m >> n$.  

\subsection{Encoding of Texts by Sequential Occurrence and by 
Rank Encoding of Terms} 
\label{textencode}

A problem we have when treating 209 (Grimm Brother) texts, characterized 
by frequency of occurrence on 7443 terms, is that we know how to generate
the vector that represents each text, but we cannot take the representation
and recreate the text.  It is a one-way encoding.  Through rank 
order encoding,  we will set up case studies so that we can go in either 
direction, between representation and object.  

For the present we have been using a contingency table of dimensions 
$209 \times 7443$ to characterize the 209 texts in the 7443-dimensional word
space.  For convenience let us take all texts to be of the same length, $L$,
so this is constant for the text set, and this can be arranged by padding
texts to a common maximum length.  The term set, of size $m$, is constant for the
text set by design. We can sparsely encode the texts, allowing 
reconstruction of each text from its code, through use of a 
$L m$-length representation vector for each text, with each value being 
either 0 or 1.  Therefore a 
text is represented in the space $\{ 0, 1 \}^{L m}$, or is a hypercube vertex
in $\R^{L m}$.   With the longest text being 8556 words, providing a value for 
$L$, and $m = 7443$, this encoding is quite impractical.  

A more economic encoding based on ranks of terms is as follows.  
For the boolean (i.e., presence/absence) encoding of our
input data, it is easy to see that integer coding is feasible, based on 
rank order of the terms used.  The rank order is one possibility among many
consistent labelings of the terms.  In our coding so far, each word in our 
text is mapped onto a boolean-valued $m$-length vector, where $m$ is our
total number of terms.  If a given word is equal to the $r$th ranked 
term, assuming terms ordered by decreasing frequency of occurrence, 
and lexicographically, then 
the $r$ location of this boolean-valued $m$-length vector has a value of 1, 
and all other locations have a value of 0.  

Using the ranks of terms occurring in texts is very straightforward.  Instead 
of the $Lm$-length representation, we get an $L$-length representation, in 
the space $\Z^L_{m+1}$.
One 
further issue must be addressed, however, and that is the varying lengths
of the texts.  The boolean encoding, above, 
used the longest text among those considered, viz.\ $L$.  Now using rank
order of terms, a simple way to make all text lengths the same is to repeat
each term in the text the requisite number of times, dropping such repetitions
for the very last term, in such a way that the overall text length in all 
cases is $L$.  With very few cases of information loss (e.g., 
``this is one very,
very, very problematic example'') the algorithm for deleting these
repeated, redundant terms is straightforward in computation (linear) and 
precision (exact recovery).    Due to semantic reasonableness we prefer 
this approach to simply padding a text to length $L$. 

To summarize, our rank ordering procedure is as follows.  The rank orders of
each term in the set of terms in our text are determined.  We will take the
rank orders as 1 = most frequent term, 2 = next most frequent term, and so on, 
through to the least frequent term.  Where terms are {\em ex aequo}, we 
use lexicographical order.  Then we replace the text with the ranks of terms.
So we have a particular, numerical (integer) encoding of the text as a whole.
For convenience we ignore punctuation and whitespace although we could well 
consider these.  In general we ignore upper and lower case.  We 
do not use stemming or other processing.  

\subsection{Haar Wavelet Transform Algorithm using Lifting}
\label{sect43}

Let's say now that we have (integer-valued) ranks, and we hierarchically 
structure objects (here, Grimm texts) that
are characterized using such data, and then we carry out our wavelet transform 
in order to have our approach to reconstructing each of the objects.  We run into 
an immediate problem if the wavelet transformed data is non-integer, and cannot
be assimilated to ranks.   We avoid this, and from 
integer input always remain with integer values, by using the lifting scheme 
\cite{sweldens} 
algorithm for the Haar wavelet transform.  

The traditional Haar wavelet transform algorithm is as follows.  From two
elements (vectors or scalars), $a, b$, we form
$ s = (a + b)/2$, and then $+d = (a + b)/2 - a = (b - a)/2$, and
similarly $-d = (a + b)/2 - b = (a - b)/2$.   Reconstructing,
$a = s - d = (a + b)/2 + (a - b)/2$, and $b = s + d = (a + b)/2 +
(b - a)/2$.

Instead, now, let $s = (a + b)/2$ as before, but take $+d = b - a$, and
$-d = a - b$.  Then reconstruction is $a = s - d/2$, and $b = s + d/2$.  We
have just let the reconstruction ``carry the burden'' of the division by 2.

The advantage of the latter procedure, referred to as the ``lifting''
algorithm, with ``Predict'' (calculating detail, $d$) and ``Update'' 
(calculation of smooth, $s$), 
is that what we store for the detail, $d$, is an integer if both
$a$ and $b$ are integers.

\begin{table}
\begin{center}
\begin{tabular}{llll}\hline
Algorithm   & Smooth   &  Detail  &  Reconstruction  \\ \hline
           &          &          &                  \\
Basic      &  $ s = \frac{a + b}{2} $
           & $+d = \frac{a + b}{2} - a = \frac{b - a}{2}$ 
           & $ a = s - d $ \\  
           &  
           & $-d = \frac{a + b}{2} - b = \frac{a - b}{2}$ 
           & $ b = s + d $ \\
           &  & &              \\
Lifting-1  &  $ s = \frac{a + b}{2} $
           & $+d = b - a $ 
           & $ a = s - \frac{d}{2} $ \\  
           &  
           & $-d = a - b$ 
           & $ b = s + \frac{d}{2} $ \\
           &  & &                         \\
Lifting-2  &  $ s = a + b $
           & $+d = b - a $ 
           & $ a = \frac{s}{2} - \frac{d}{2} $ \\  
           &  
           & $-d = a - b$ 
           & $ b = \frac{s}{2} + \frac{d}{2} $  \\
           &  & &                                   \\ \hline
\end{tabular}
\end{center}
\caption{Haar wavelet transform schemes.}
\label{tabhaar}
\end{table}

\subsection{Integrated Rank-Based Representation and an Example}

\begin{figure*}
\begin{center}
\includegraphics[width=14cm]{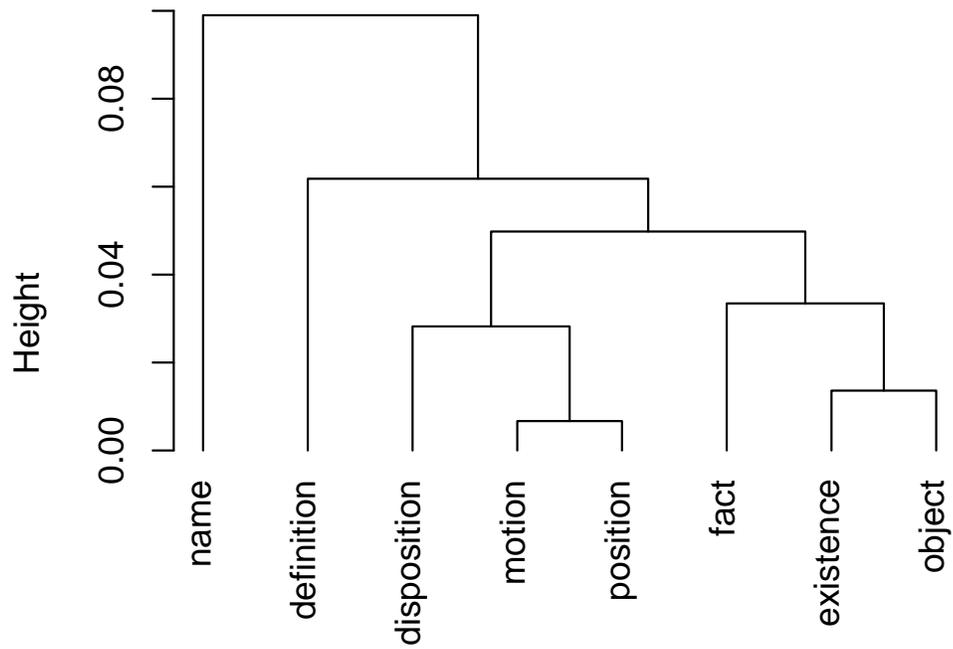}
\end{center}
\caption{Hierarchical clustering of 8 terms.  Data on which this 
was based: frequencies of occurrence of 66 nouns in 24 successive, 
non-overlapping segments of Aristotle's {\em Categories}.}
\label{arist-ranks-1}
\end{figure*}
\begin{figure*}
\begin{center}
\includegraphics[width=14cm]{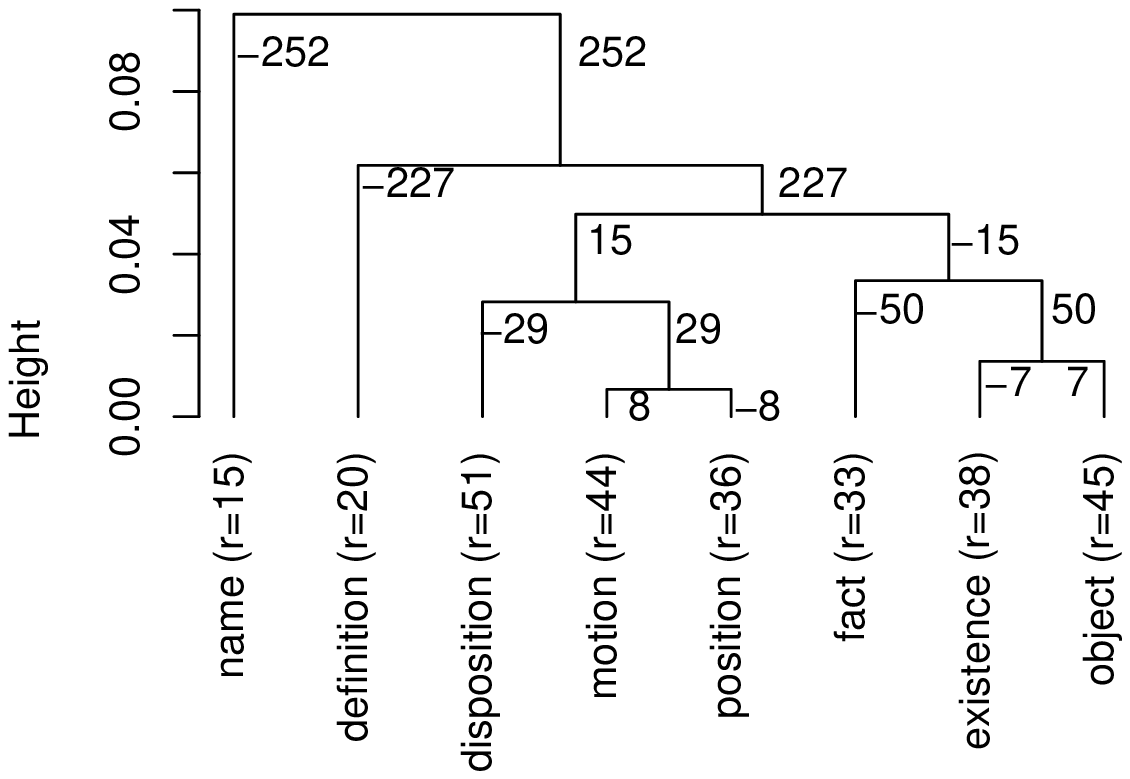}
\end{center}
\caption{Haar dendrogram wavelet details for the hierarchy corresponding 
to Figure \ref{arist-ranks-1}.  The wavelet transform is based on the ranks
of the words, which are shown in the word labels.}
\label{arist-ranks-2}
\end{figure*}

Let $r$(word) be the rank of a word.  We have that:

\begin{itemize}
\item 
We can determine the smooth of word$_i$ and word$_j$, through use of 
$r$(word$_i$) and $r$(word$_j$).  In line with the Lifting-2 scheme in 
Table \ref{tabhaar}, the smooth is $r$(word$_i$) + $r$(word$_j$).
\item 
The detail signal then is $\pm | r$(word$_i$) $ - r$(word$_j) | $
\item Furthermore there is some word$_k$ such that 
$r$(word$_k) = | r$(word$_i)  - r$(word$_j) | $ so that a detail 
coefficient is given by $\pm $ word$_k$ for some $k$.  
\item In fact, when  $\pm $ word$_k$ is the detail, we have the 
following linear relationship: \\
$2 \cdot$ word$_i$ = smooth(word$_i$, word$_j$) $ - $ word$_k$ \\
$2 \cdot$ word$_j$ = smooth(word$_i$, word$_j$) $ + $ word$_k$ 
\item Finally it is likely that word$_k$ is not in the word set that we are 
examining.  We adopt an easy solution to how we represent word$_k$ through
its rank, $r$(word$_k$).  Firstly, word$_k$ can be from a superset of the
word set being analyzed; and we allow multiples of our top rank to help with 
this representation.   Figures, to be discussed now (Figures 
\ref{arist-ranks-1} and \ref{arist-ranks-2}), will exemplify this.
\end{itemize}

Our aim is to have a closed system where the dendrogram wavelet transform
of words transforms to words. 
We will illustrate this generally applicable procedure
 using an example.  We took Aristotle's 
{\em Categories}, which consisted of 14,483 individual words.  
For expository purposes, as we will now see, we selected a small subset of 
words.  

The procedure followed, with motivation, is as follows.  We broke 
the text into
24 files, in order to base the textual analysis on the sequential properties
of the argument developed.  In these 24 files there were 1269 unique words.
We selected 66 nouns of particular interest.  A sample (with 
frequencies of occurrence) follows:
man (104), contrary (72), same (71), subject (60), substance (58), ...
No stemming or other preprocessing was applied.  

The first phase of processing is to construct a hierarchical 
clustering. 

For the hierarchical
clustering, we further restricted the set of nouns to just 8.  (These will
be seen in the figures to be discussed below.)  
The data array was 
{\em doubled} \cite{murtaghca} to produce an $8 \times 48$ array, which with 
removing 0-valued text segments (since, in one text segment,
 none of our selected 8 nouns appeared) gave an $8 \times 46$ array,
thereby 
enforcing equal weighting of (equal masses for) the nouns. 
The spaces of the 8 nouns, and of the 23 text segments (together with the
complements of the 23 text segments, on account of the data doubling) 
are characterized prior to the correspondence analysis in terms of their
frequencies of occurrence, on which the $\chi^2$ metric is used.  The 
correspondence analysis then ``euclideanizes'' both nouns and text segments.
Such a Euclidean embedding is far safer for later processing, including 
clustering (and frankly would be most ad hoc, and/or ``customized'' and 
less general, in terms of any alternative data analysis).  We used a 
7-dimensional (corresponding to the number of non-zero eigenvalues)
Euclidean embedding, furnished by the projections onto the 
factors.  A hierarchical clustering of the 8 nouns, characterized by their
7-dimensional (Euclidean) factor projections, was carried out: Figure 
\ref{arist-ranks-1}.  The Ward minimum variance agglomerative criterion 
was used, with equal weighting of the 8 nouns.  

Based on the hierarchical clustering, the second phase of processing 
is to carry out the wavelet transform on it.  

Using the ``Lifting-2'' scheme in Table \ref{tabhaar} ensures that 
Haar dendrogram wavelet detail and smooth components will be integers which 
we can read off as ranks.  The result of this processing is shown in 
Figure \ref{arist-ranks-2}.  The ranks are used here, and are noted in 
the terminal labels.  The wavelet transform, using the ``key'' of the
hierarchy, is based on these ranks.  
The overall smooth (not shown) at the root of 
the tree is 282.  So, $s_7 = 282$.  We use the ``Lifting-2'' scheme in 
order to ensure that integers, that we will interpret as ranks, are used
as details and smooths throughout.  

Let's check how we reconstruct, say, ``disposition''.  From Table
\ref{tabhaar}, we have, where, as we have noted,
 the final smooth, 282, is not shown in Figure \ref{arist-ranks-2}:
$ 
(((((282 + 252)/2 + 227)/2 + 15)/2 -29)/2) = 51$.
We have therefore traced the path from root to the terminal node corresponding
to ``disposition'', accumulating final smooth and details, and carrying out
the division by 2 as per Table \ref{tabhaar}.

Our next step is to give a meaning to the details, and final smooth, based
on the word set used.  But we have used 66 words in all.  Let us therefore 
define ranks 227  =   3* 66 + 29; 252  =   3* 66 + 54; and 282  =   4* 66 + 18.
Next, we check what words we in fact have for the ranks that we use here:
$\{66, 18, 54, 29, 15, 50, 29, 7, 8 \}$ = $\{$
 ``number'',  ``parts'',  ``affections'',  ``sense'', ``name'',  
``correlatives'',  ``sense'', ``knowledge'', ``qualities'' $\}$.

Now, back to reading off the trajectory of ``disposition''.  We can 
rephrase this in terms of words:

``disposition'' = 
((((( 4 * ``number'' + ``parts'' + 3 * ``number'' + ``affections'')/2 
+ 3 * ``number'' + ``sense'')/2 + ``name'')/2 $-$ ``sense'')/2)

So we can say that ``disposition'' is a simple linear combination 
 of the following terms:
``number'', ``parts'', ``affections'', and ``sense''.

Having shown that we can define a word in terms of other words,
we can carry out  the same calculation for all others here.  

Let us note that, by using the entire hierarchy of embedded sets, a
very simple alternative expression is available for any individual 
concept.  Label the non-terminal nodes as follows: $n_1$ = $\{$ ``motion'', 
``position'' $\}$; $n_2$ = $\{$ ``existence'', ``object'' $\}$; $n_3$ = 
$\{$ ``motion'', ``position'', ``disposition'' $\}$; etc.  Then ``name'' = 
$n_7 - n_6$.  ``Definition'' = $n_6 - n_5$.  We can continue straightforwardly
to label concepts on this basis.

In the representation used for our selected set of words from the 
Aristotle text, let us note that this representation is an integer one (i.e., 
ranks, which in the dendrogram wavelet transform, are processed as integers).
One important conclusion we draw is that with such a representation we 
must represent wavelet details and the smooth in terms of words that 
are not in the selected set.  We must go beyond the selected set.  
In our Chernoff face case study, section \ref{sect6},
we used the reals in our representation, 
and then the details and the smooth came from the same set (i.e., continuous 
interval).  

\subsection{Wavelet Coefficients Derived From a Rank Encoding}

The wavelet transform of rank data has interesting links between 
details and rank correlation; and smooth and rank concordance.

Spearman's coefficient $\rho$ of rank correlation is defined from 
the squared differences of ranks.  
(Take two rankings, $\{ r_j, r'_{j} | 1 \leq j \leq n \}$.  Then 
$\rho = 1 - 6 \left( \sum_{1 \leq j \leq n} 
(r_j - r'_j)^2 \right) / (n^3 - n)$.  See \cite{kendall}.)  Using the 
lifting scheme, the detail coefficients
in the Haar wavelet transform are differences of ranks.  The energy of 
the  detail coefficients can be 
viewed therefore as contributions to a Spearman's $\rho$.   Kendall 
(\cite{kendall}, chapter 9), deals with 
approximations for $\rho$ for $n$ large, and close relationship between 
use of ranks and 
 of equivalent real-valued variates.   

To provide an interpretation for such rank correlations, consider the case 
of a perfectly balanced or regular
hierarchy.  Then the first tranche of detail values will be between 
successive pairs of our input vectors.  
The contributions to Spearman's $\rho$ are between these pairs.  For 
the next tranche of nodes in the 
hierarchy, we are considering successive pairs of non-terminal nodes, 
with the implication that our 
contributions to Spearman's $\rho$ deals with correlation between these 
pairs.  Ultimately, therefore,
the contributions to Spearman's $\rho$ correlation are between successive 
pairs, of input vectors, and 
of clusters of them, read off in accordance with the sequence of 
agglomerations in the hierarchy.  

The overall or final smooth is based on repeatedly summing 
 ranks.  The average ranking 
is used  in the concordance of the set of rankings, Spearman's coefficient 
of concordance (see chapters 6, 7
of \cite{kendall}).  In fact, 
treating the ranks as real-valued variates can allow us, ``with caution''
(\cite{kendall}, p. 125) to 
arrive at the same outcome as if we had used real-valued variates to start 
with.  

\section{Complexity of an Object}
\label{sect55}

In the context of $n$ texts, and the earlier face case study,
we have considered the following, where $m$ is the number of unique 
attributes (words; face attributes), and $L$ is the maximum object
(text, face) size or total number (non-unique) of attributes.

\begin{itemize}
\item A ``bag of words'' description of a given text, leading to an $m$-length 
representation.  Directly generating one text requires $2^m$ decisions or 
operations.  
A random, assuming uniformity, text has probability $2^{-m}$.  A Shannon 
information measure of the object is $m$ bits. 

\item A boolean description of a given text, with an $Lm$-length representation 
for each text.  Directly generating one text requires $2^{Lm}$ operations.
A random, assuming uniformity, text has probability $2^{-Lm}$.  A Shannon 
information measure of the object is $Lm$ bits. 

\item A rank description of a given text, using ranks $1, 2, \dots, R$.  Each 
text has an $L$-length represenation.  
Directly generating one text requires $R^{L}$ operations.
A random, assuming uniformity, text has probability $R^{-L}$.  A Shannon 
information measure of the object is $CL$ bits, where $C$ is a constant. 

\item In the face case study, the representation was $m$-length and real.  
Let each real value be discretized into $P$ intervals.  Each face then 
is described by a boolean $Pm$-length representation.  
Directly generating one face requires $2^{Pm}$ operations.
A random, assuming uniformity, face has probability $2^{-Pm}$.  A Shannon 
information measure of the object is $Pm$ bits. 

\end{itemize}

Now we change the context, and assume that we have a hierarchical structuring 
of the set of $n$ objects considered.  Directly generating one object requires 
$O(n)$ operations, and worst case $n-1$.  Each operation is of linear 
computational complexity in the representation used.  
A random, assuming uniformity, object has probability $n^{-1}$.  A Shannon
information measure of the object is $\log n$ bits.

Our interest lies in cases where $n << m, L, P$, i.e.\ the total number of 
objects considered is very much less than the length of their description. 

In practice, given a natural macroscopic object class, $n$ may be small, 
whereas we can go to great lengths to characterize the objects in terms
of precision or description length.
So the computability of the object is likely to be far more tractable,
given our approach based on the hierarchical coding of information.

\section{Conclusions}
\label{sect66}

Our approach has been inspired by algorithmic information (or Kolmogorov
complexity) that considers a single, finite object and, more particularly,
the length of the shortest binary program from which the object can be 
effectively reconstructed.  As a tool, we used a novel wavelet transform 
on a hierarchy to provide a layer-by-layer reconstruction of the object,
starting from an average object (under certain circumstances, a mean 
object).  

Significant challenges are facing us in regard to how we understand and 
process objects, as noted by Brooks \cite{brooks}.  A solution that we
propose from the work described in this article is  
to explore further ``hierarchical coding systems'' (this characterization is
used in \cite{khrenn}) of the 
sort used in this work.  We
have described in this work how this can be done, using a range of 
 practical, simplified, case studies.  

We have shown, theoretically and in case studies, that we can: 

\begin{itemize}
\item generate faces from faces, with a global sum (or average) 
face as our starting 
point, 
\item generate concepts from concepts, with an average concept as our 
starting point.
\item We have discussed how we can go further, to deal with, say, a document
space.
\end{itemize}

Our generation procedure is of average complexity proportional to $\log n$, and 
worst case $O(n)$, when we are dealing with $n$ objects (faces, concepts,
etc.).

Our work is consistent with \cite{hawkins}, who in a machine learning
perspective, concludes that:
``We have recognized a fundamental concept of how the neocortex uses
hierarchy and time to create a model of the world and to perceive
novel patterns as part of that model.''

Anderson \cite{anderson2} remarks on how ``it may be the case that the
unique reach and power of human ... intelligence is a result not so
much of a unique ability to perform complex, symbolic cognition in
abstraction from the environment, but is rather due in large measure
to the remarkable richness of the environment in which we do our thinking.''
He elaborates on this as follows.
A central role is played ``by persisting institutions and practices in
supporting the possibility of high-level cognition.  In cognitive science
such structures are called scaffolds; a scaffold, in this sense, occurs
when an epistemic action results in some more permanent cognitive aid --
symbolic, or social-institutional.''  So ``we do very complex things,
e.g., building a jumbo jet or running a country 'only indirectly -- by
creating larger external structures, both physical and social, which can
then prompt and coordinate a long sequence of  individually tractable
episodes of problem solving, preserving and transmitting partial solutions
along the way' \cite{clark}.  These structures include language, especially
written language, and indeed all physically instantiated representations or
cognitive aids ... Such scaffolds allow us to break down a complex problem
into a series of small, easy ones, ... Not just symbol systems, but
social structures and procedures can sometimes fill a similar role.''

All of this is exciting, but it rests on a fundamental bedrock of 
representation in the sense of data encoding, together with composition 
operators defined on these codes.  We require, as a {\em sine
qua non} for this work, a data encoding scheme (i) preferably of small, finite
length, (ii) capable of being efficiently (low order polynomial) converted
into a display, and (iii) capable of being efficiently (low order polynomial)
determined from a real world exemplar of the object.  

Mainstream physics proceeds by analyzing the ever smaller and ever larger.  
Mainstream computer science has its point of departure in the necessary 
finiteness of that which is computed.  The feasibility of this computer science
perspective is based on our finiteness as human beings.  An interesting 
example from \cite{dix}, discussed in \cite{ohara}, is to consider a person 
monitored by a video camera for their entire life.  The amount of data, 
for 70 years or $2.2 \times 10^9$ seconds, is to an approximation 
27.5 terabytes.  Let us pose the question of the complexity of a human 
life, expressed as this particular 27.5 terabytes of information.  
In a similar vein, the work of Shakespeare, according to \cite{buckley},
amounts to under one million words, and can be spoken in 70 hours. 


A further supporting view, for music and literary works, is as follows.
Basing himself approvingly on a publication by R. Kolisch in 1943, 
musical and cultural theorist Adorno \cite{adorno}
considered ``the basic characters to which the types of Beethoven's
tempi correspond.  In this way, [we arrive] at a discrete number
of such basic characters and tempi.  At first, the result is shocking;
it seems a bit mechanistic and overly mathematical in relation to 
Beethoven's gigantic oeuvre.  But if you turn the tables, ... you 
will find that great ... music actually bears some resemblance to 
a puzzle.  The movements of the greatest composers are based on a 
discrete number of {\em topoi}, of more or less rigid elements,
out of which they are constructed.  ... Music represents itself as 
if one thing were developing out of the other, but without any such
development literally occurring.  The mechanical aspect is covered 
up by the art of composition, ...''.  Adorno's discussion continues 
with a reference to a similar picture in relation to how ``Similarly,
with a certain amount of na\"{\i}vit\'e, the great
philosophical systems beginning with Plato have had recourse again and 
again to such mechanical means ...''.  

Our perspective, based on some hierarchically structured, appropriate 
representation or encoding of our object family, and an associated
algebra, is that it is so much 
easier to grow the object!  Algorithmic complexity traditionally is 
related to the length (or size) of the object.  For us, algorithmic 
complexity is related to the size of the object class, rather than 
to the size of the object.  Such a perspective
is not a replacement for the algorithmic information view.  It is simply 
a different view.  

In physics,
the pursuit of the ever smaller and ever larger, notwithstanding 
finite and discrete limits,  make the computability of physical objects
difficult and problematic.  On the other hand the finitary computer science
view presented in this work, based on hierarchical coding,
is eminently tractable and allows natural and 
artifact objects to be computable.  


\end{document}